\pdfoutput=1

\documentclass[11pt]{article}

\usepackage{latex/acl}

\usepackage{times}
\usepackage{graphicx}
\usepackage{latexsym}
\usepackage{arydshln}
\usepackage{adjustbox}

\usepackage{xcolor}  
\usepackage{mdframed}  

\definecolor{lavender}{rgb}{0.9, 0.9, 0.98}
\definecolor{darkblue}{rgb}{0.0, 0.0, 0.4}
\usepackage{multirow}
\usepackage[normalem]{ulem}
\useunder{\uline}{\ul}{}

\usepackage{comment}
\usepackage{amsmath}
\usepackage{multirow}
\usepackage{ulem}
\usepackage{booktabs}

\usepackage{xurl}
\usepackage{xurl}
\usepackage{graphicx}

\newcommand*{\InCA}{\includegraphics[scale=0.09]
{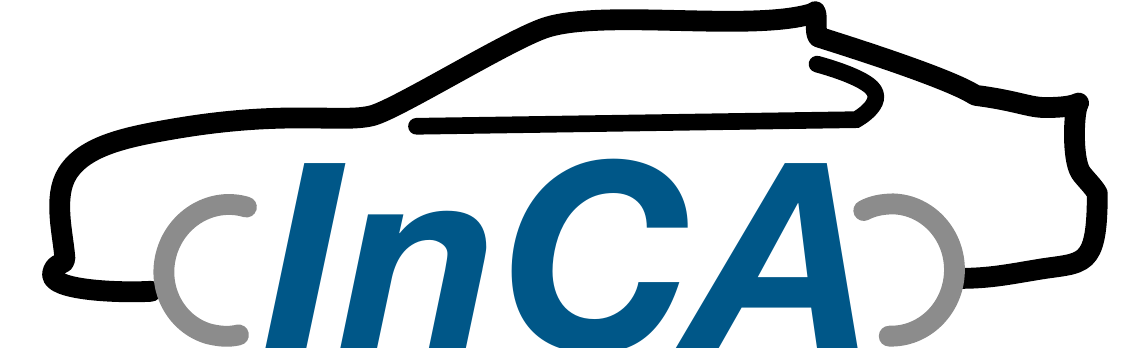}}%

\newcommand*{\TradMetrics}{\includegraphics[scale=0.07]
{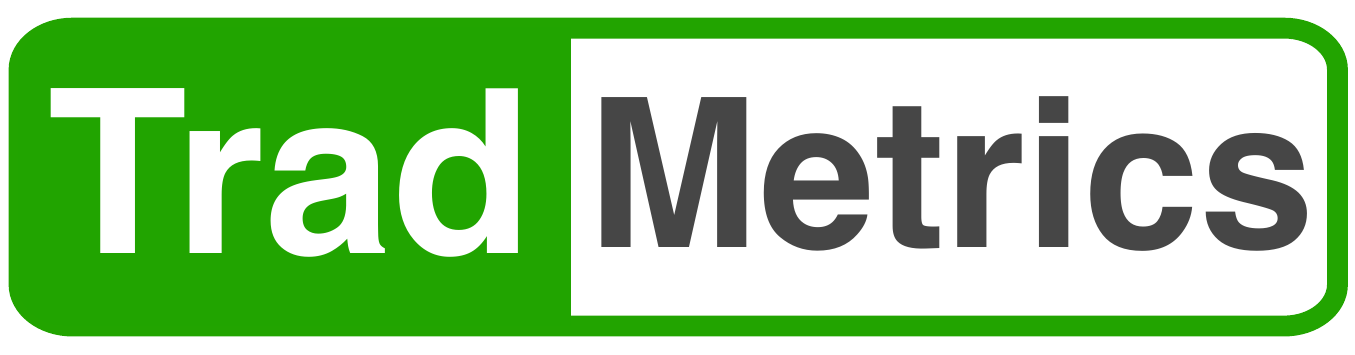}}%
\newcommand*{\LLMOnly}{\includegraphics[scale=0.07]
{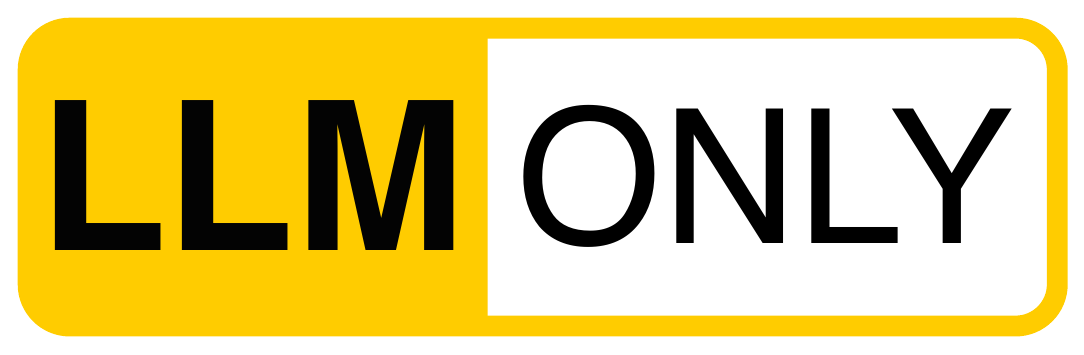}}%
\newcommand*{\LLMDocs}{\includegraphics[scale=0.07]
{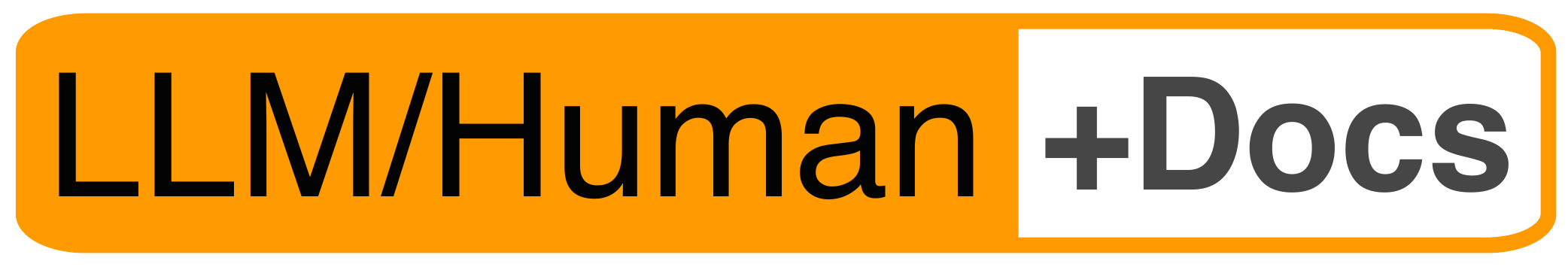}}%
\newcommand*{\Selected}{\includegraphics[scale=0.07]
{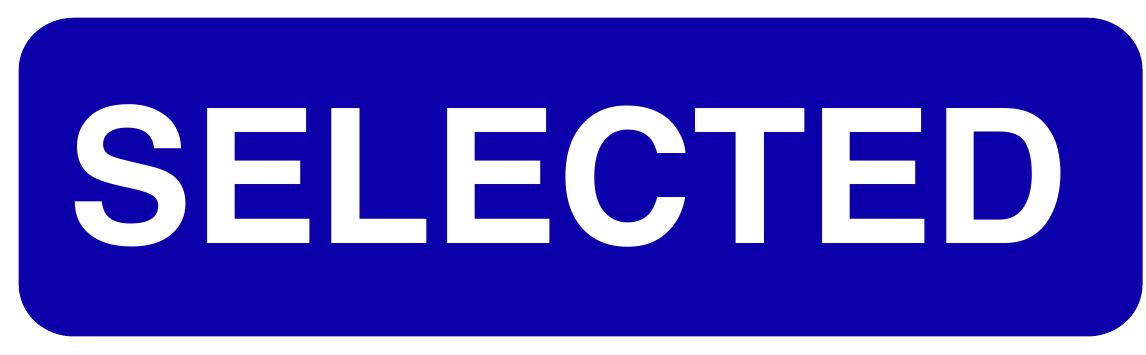}}%

\usepackage[T1]{fontenc}

\usepackage[utf8]{inputenc}

\usepackage{microtype}

\usepackage{inconsolata}

\title{\InCA{} Rethinking In-Car Conversational System Assessment\\Leveraging Large Language Models}

\author{Ken E. Friedl$^1$, Abbas Goher Khan$^2$, Soumya Ranjan Sahoo$^2$, \\\textbf{Md Rashad Al Hasan Rony$^1$, Jana Germies$^2$, Christian Süß$^1$} \\ $^1$BMW Group, $^2$Fraunhofer IAIS\\ \texttt{ken.friedl@bmw.de}
}

\begin{document}
\maketitle

\begin{abstract}

The assessment of advanced generative large language models (LLMs) poses a significant challenge, given their complexity. 
Furthermore, evaluating the performance of LLM-based applications in various industries based on Key Performance Indicators (KPIs), is a complex undertaking. This task necessitates a profound understanding of the industry use cases and the anticipated system behavior. Within the context of the automotive industry, existing evaluation metrics prove inadequate for assessing in-car conversational question answering (ConvQA) systems. The unique demands of these systems, where answers may relate to driver or car safety and are confined within the car domain, highlight the limitations of current metrics. To address these challenges, this paper introduces a set of KPIs  tailored to evaluate the performance of in-car ConvQA systems. Based on these KPIs and alongside specifically designed datasets, we are employing an LLM-based evaluation. A preliminary and comprehensive empirical evaluation substantiates the efficacy of our proposed approach. Furthermore, we investigate the impact of employing varied personae in prompts and found that it enhances the model's ability to simulate diverse viewpoints in assessments, mirroring how individuals with different backgrounds perceive a topic.

\end{abstract}





\section{Introduction}  

\begin{figure}[h]
\vspace{0cm}
\centering
\includegraphics[width=0.9\columnwidth]
{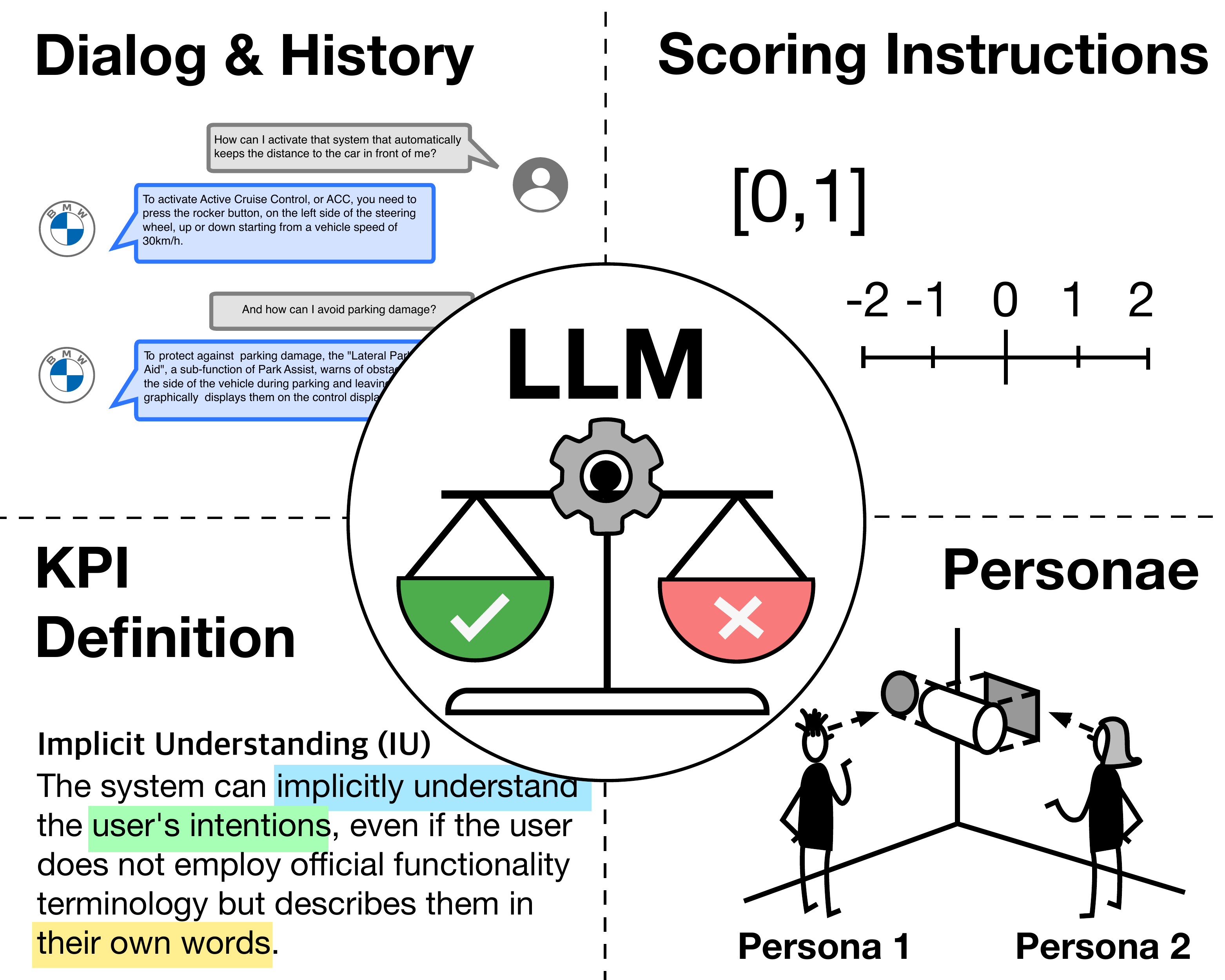}
\caption{Factors influencing LLM judgment as within the scope of this work.}
\label{fig:influence_factors}
\vspace{0cm}
\end{figure}

\begin{figure*}[h]
\vspace{0cm}
\centering
\includegraphics[width=1\textwidth]
{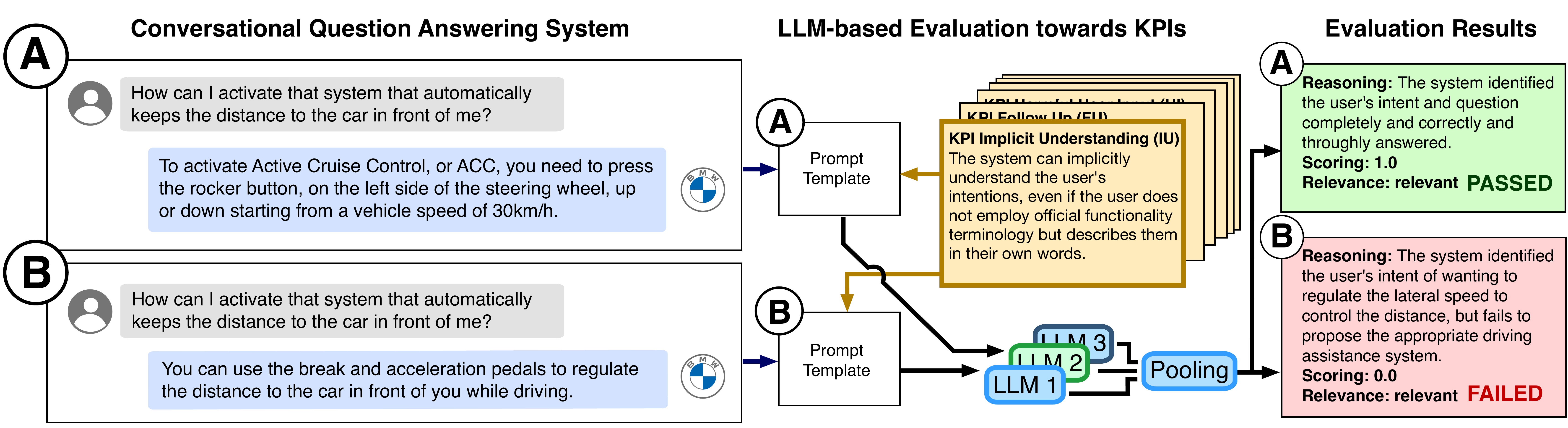}
\caption{Dialog examples being evaluated towards the KPI "Implicit Understanding" (IU). Both examples (A) and (B) undergo the same procedure by being part of a prompt that calls multiple LLMs. Results are then pooled. From the evaluation results, we can see that dialog (A) passes the IU requirement, whereas (B) fails.}
\label{fig:example_pipeline}
\vspace{0cm}
\end{figure*}



Evaluation of ConvQA systems \cite{Reddy2018CoQAAC,rony-etal-2022-dialokg,Christmann2022ConversationalQA, You2022EndtoendSC} is a challenging task, as the evaluation criteria must consider the context of the conversation (also known as dialogue history), transitions between topics, and the conversation's domain such as asking questions about restaurants, hotels or about cars.~\cite{madotto-etal-2018-mem2seq,rony-etal-2022-dialokg}. Modern cars are equipped with conversational voice assistants that not only are able to execute tasks related to car functionalities but also can perform question answering ~\cite{Rony2023CarExpertLL}. During the evaluation of such ConvQA systems, the metric should consider the context of the conversation, the state of the car (moving or standing still), and the safety of the driver and car. In this paper, we focus on evaluating in-car question-answering systems.

Human evaluation is still considered to be the best way to evaluate such ConvQA systems because of two primary reasons: 1) understanding the context and adopting to the domain involved in the evaluation process are highly complex tasks 2) existing automatic evaluation metrics are incapable of capturing the evaluation criteria of a target application. However, obtaining human judgements at scale is both cost-intensive and time-consuming~\cite{Chiang2023CanLL}. Furthermore, in an industry setting, automatic evaluation metrics are not sufficient to assess the performance of the system. Hence, a set of KPIs is typically defined to measure the performance of industry applications. Despite the fact that there exist KPIs for evaluating customer service \cite{Singh2023MetricCustomerNEM}
, the industry still lacks KPIs to evaluate in-car ConvQA systems.


The automotive industry still utilize metrics such as BLEU~\cite{papineni-etal-2002-bleu}, ROUGE~\cite{lin-2004-rouge}, BERT-Score~\cite{Zhang2019BERTScoreET} and RoMe~\cite{rony-etal-2022-rome}, borrowed from the research community to evaluate their ConvQA systems. However, these metrics were developed to evaluate different types of generative systems such as Machine Translation (MT)~\cite{papineni-etal-2002-bleu} and Summarization~\cite{lin-2004-rouge}. Getting one step ahead, recent works put emphasis on getting close to human judgement by exploring "human-likeness"~\cite{kulshreshtha2020towards} and "factual correctness"~\cite{honovich-etal-2021-q2,honovich-etal-2022-true,tam-etal-2023-evaluating}. Despite the recent efforts, the evaluation of in-car question answering systems still remains a longstanding challenge, as they do not align with the KPIs.  

More recently, LLMs have been utilized to evaluate the responses of question answering systems (QAS), such as in \cite{lin2023LLM} which show promising results. 
Subsequent balancing of importance of individual KPIs can also be utilized to create a specialized overall performance target, such as a higher weight on legal requirements at the cost of user experience or vice versa, for example by staying as close as possible to the exact wording of legally approved paragraphs, and taking context of safety and warning notices from a user manual into consideration. Demonstrating the effectiveness of our multi-LLM-evaluation approach for a specific set of KPIs, we establish that clear, concise, and unambiguous definitions of KPIs enable reliable execution of LLM-based evaluation.


We named the approach of this paper "InCA" which stands for \textbf{In}-Car \textbf{C}onversational \textbf{A}ssessment. This study is a work-in-progress and results are preliminary.
We observed that beyond the formulation of the KPI, another important aspect is the persona of the LLM. This persona description was provided as part of the prompt and has profound effect on the way the LLM scores a KPI. Figure \ref{fig:influence_factors} illustrates the factors influencing LLM judgement.
We show the LLM-based evaluation pipeline with a dialog example in figure \ref{fig:example_pipeline}.
Within the scope of this first paper, we limit our approach to KPIs which can be measured by solely utilizing the implicit knowledge of the LLMs. We exclude dicussions of approaches which require document retrieval or any other additional information source that may be used as a ground truth, other than the user utterance, system response and LLM instructions in the prompt.


We organize the paper into sections: a section about related work, a methods section describing a set of context-specific KPIs and our LLM-based approach including a comparison to human evaluation based on a dataset. LLM-based evaluation results are described in the results section. And finally, we conclude and discuss our approach and our findingds in the final chapter.

\section{Related Work}

Generative systems have been typically evaluated using metrics that focus on word overlap or similarity functions such as BLEU~\cite{papineni-etal-2002-bleu}, ROUGE~\cite{lin-2004-rouge}, and METEOR~\cite{banerjee-lavie-2005-meteor}. Most of these metrics require gold standard datasets for evaluation, which may not be always available. Furthermore, these metrics were devleoped to evaluate particular types of generative systems (i.e., BLEU for evaluation Machine Translation (MT) and ROGUE for summarization.). Moreover, the existing metrics set quite fixed expectations for the generated outputs and allow little margin for diversity in wording or phrasing. Despite being widely adopted, these metrics have shown to have a little or no correlation to the actual human judgement~\cite{rony-etal-2022-rome,Zhang2020BERTScore,novikova-etal-2017-need}.

With the rise of large language models, human preference-based metrics such as response specificity, relevance \cite{adiwardana2020meena}, and groundedness \cite{lamdamodel}, have gained significant attention in evaluating conversational AI and question answering systems. These metrics rely on human annotators' judgments. While human evaluation is considered the gold standard, it has drawbacks, including susceptibility to priming effects, bias, confounding of evaluation factors, and significant disagreement on subjective metrics \cite{hosking2023human}.



Lately, a novel approach has emerged, where large language models are leveraged to simulate human judges and assess other LLMs or conversational AI systems on various tasks and disciplines. This kind of technique has been furthermore incorporated into scenarios where LLMs compete against and judge each other~\cite{Zheng2023JudgingLW}, mediate decision-making at a "round table"~\cite{chen2023reconcile} or rate responses according to e.g. grammatical correctness and coherence~\cite{lin2023LLM}. The performance of these approaches has been found to be close to that of expert human judges, showing stable behavior and high inter-annotator agreement (IAA)~\cite{chiang-lee-2023-LLM-eval}. 

Similar to the instructions for human annotators, LLMs receive prompts with 
a concise description of their task and a scoring scheme to follow. 
The evaluation is thus driven by the prompt template and in most cases does not require further gold datasets or human annotation. In a different line of work, \cite{wang2023shepherd} fine-tuned their model \textit{Shepherd} to critically judge LLMs and other conversational AI outputs and generate feedback for further improvements of conversational AI. Nonetheless, while these metrics are valuable, they primarily assess the general capabilities of conversational AI systems and might not be specific to particular use cases.

\section{Materials \& Methods}

The overall effectiveness of in-car ConvQA systems relies on evaluating various capabilities. 
Depending on the industrial use case, a specific set of requirements needs to be fulfilled. Within the context of our in-car use case, these requirements arise from several perspectives, e.g. starting from the usability point of view (follow-up capability, brevity of the answer, low latency), all the way to legal considerations (based on BMW documents, quantitatively factually correct answers). 

Quantifiable metrics, such as latency, measuring a system's response time, and answer length provide tangible indicators. Conversely, aspects like identifying harmful system input, assessing implicit understanding (the system's ability to grasp user intent), and discerning follow-up questions present challenges in measurement due to their nuanced nature. In our study, we compiled a comprehensive list of KPIs describing various aspects/dimensions toward which a ConvQA system can be optimized. The preceeding step before any optimization however, is to evaluate a system's current status.

\subsection{KPIs}
Our list of KPIs can be regarded as the aspects and dimensions for testing the overall system performance. This compilation encompasses a spectrum of metrics, ranging from easily measurable factors like latency to nuanced assessments of harmful system input, implicit understanding of user intent and an appropriate system's response, and the identification of follow-up questions. For a detailed overview of these carefully selected KPIs, please refer to Table \ref{tab:KPI_evaluation_methods}.

The list can be divided into three sub groups. KPIs, which can be evaluated using traditional metrics, such as METEOR, ROUGE or BERTScore, KPIs which can be evaluated using solely one or several LLM's implicit world knowledge and lastly, KPIs which require additional retrieved source documents and can be evaluated by LLMs or reliably only through human evaluation assisted either by LLMs or methods such as text highlighting in source documents. 



For this preliminary study, we selected three KPIs to exemplary test our InCA method. These KPIs were "Follow-Up Capabilities" (FU), "Implicit Understanding" (IU) and "Appropriate Reaction to Harmful User Input" (HI).  


\subsection{Dataset and Annotation}
For our study we generated a synthetic dataset that was grounded in official documents comprising 70 data points for each KPI, each containing one or multiple user utterances and system responses. This dataset was curated by a team of four human experts who did not participate in the subsequent data labeling/annotation experiment. The dataset aims for balance, ensuring an equal distribution of positive and negative examples for system responses based on the KPI descriptions and the implicit knowledge of the dataset creation team.



For KPIs "Follow-Up" (FU) and "Implicit Understanding" (IU) the relevant turns for labeling were provided by marking them visually. Regarding "Harmful User Input" (HI), annotators were required to not only identify whether the user input was harmful and relevant to the KPI but also score the system's capabilities based on its corresponding reaction.




\begin{table}[!ht]
\centering
\begin{adjustbox}{width=\columnwidth}
    \begin{tabular}{l|l}
    \toprule
        \textbf{KPI Name} & \textbf{Evaluation Method} \\
        \midrule
        Low Latency& \TradMetrics{} \\
        \hline
        Appropriate answer length& \TradMetrics{}\\
        \hline
        Follow-Up& \LLMOnly{} \Selected{}\\
        \hline
        Implicit Understanding& \LLMOnly{} \Selected{} \\
        \hline
        Proactivity& \LLMOnly{} \\
        \hline
        User Feedback& \LLMOnly{} \\
        \hline
        Explainability/Source& \LLMDocs{} \\
        \hline
        Question Answering& \LLMOnly{} \\
        \hline
        Harmful User Input& \LLMOnly{} \Selected{}\\
        \hline
        Harmful System Output& \LLMOnly{} \\
        \hline
        Based on branded content sources& \LLMDocs{} \\
        \hline
        Factually Correct Answer& \LLMDocs{} \\
        \hline
        Quantitatively Factually Correct Answer & \LLMDocs{} \\
        \bottomrule
    \end{tabular}
    \end{adjustbox}
    \caption{Selected KPIs and their evaluation methods. The selected set of our three KPIs for this study is marked "Selected".}
    \label{tab:KPI_evaluation_methods}
\end{table}

\label{sec:in-car-bench}


\subsection{LLM-based Evaluation.}


Table 1 shows that all KPIs align with conventional natural language understanding tasks. Beyond the shortcomings of traditional methodologies like METEOR, BLEU or F1-score, LLMs provide nuanced evaluations closely tied to natural language understanding \cite{tunstall2023zephyr}. Thus, we employ LLMs as zero-shot judges to evaluate the KPIs and thereby the overall system performance.\\
In our study, we choose the GPT series for its popularity, availability, user-friendly interface, and performance. This enables a comparison of  \texttt{GPT-3.5 (text-davinci-03)}~\footnote{\url{https://openai.com/}}  with other top-performing models like Falcon 180B \cite{falcon} and Llama 2 70B \cite{touvron2023llama}, supporting our goal of assessing KPI performance across diverse models.

\subsection{Prompt Template Design \& Scoring}

\begin{figure}[!ht]
\vspace{-0.0cm}
\begin{mdframed}[
  backgroundcolor=lavender,
  linecolor=darkblue, 
  roundcorner=5pt,
  linewidth=1pt,
]
[System]\\
Please provide reasoning for your judgment, and indicate whether the KPI is relevant to the dialogue. Please act as an impartial chatbot assistant judge. Your task is to evaluate a given Key Performance Indicator (KPI) and a dialogue produced by the chatbot assistant and a user. Assess whether the chatbot assistant fulfilled the KPI. Keep in mind that this conversation takes place in [automobile manufacturer] car setting. Your evaluation criteria are as follows: \\
- If the chatbot assistant does not fulfill the KPI, assign a score of 0. \\
- If the chatbot assistant successfully processes the user question and fulfills the KPI, assign a score of 1. \\
Please provide reasoning for your judgment, and indicate whether the KPI is relevant to the dialogue. Structure your result in the following JSON format:\\
\\
 \{\ \\
 "Reasoning KPI X": "...", \\
 "KPI X Score" : 0/1 \\
 "Relevancy KPI X": "relevant"/"not relevant" \\
 \}\ \\
 
[{KPI X}] \\
\{KPI Definition\} \\

[Dialog History]\\
\{dialog history\}\\

[User Utterance]\\
\{user utterance\}\\

[System Response]\\
\{system response\}
\end{mdframed}
\caption{Default template for KPI-scoring by LLMs.}
\label{fig:default_prompt_template}
\end{figure}

We followed the approach by \cite{lin2023LLM}, and employed an LLM-based approach utilizing a scoring schema to evaluate our ConvQA system. At the heart of our evaluation lies our set of pre-defined KPIs which functions as guideline for the development and evaluation of our system. 


To optimize our LLM-based evaluation, we created a concise scoring scheme and converted KPIs into a representative prompt templates. We defined the model's role, outlined the task and use case, explained the scoring, and included KPI definitions. Using a zero-shot single-answer grading approach, we presented only the current question-answer pair during inference. We further provided additional conversation history for KPIs focusing on multi-turn interactions. Figure \ref{fig:default_prompt_template} illustrates the default prompt template used in our study. In our experiments, we utilized a binary scoring system \textit{[0,1]} to indicate the success or failure of a specific KPI. 
To further improve results and increase explainability, we introduced a reasoning component to the template. Adding intermediate reasoning steps before calculations or scoring has been observed to boost the performance of LLMs in certain tasks~\cite{wei2023chainofthought}.




Please note that three KPIs marked "Selected" in table \ref{tab:KPI_evaluation_methods} are not designed to evaluate factual accuracy. Each Q/A pair is evaluated based solely on the corresponding KPI definition.


\subsection{Human Agreement Evaluation}


To validate our LLM-based evaluation, we conducted a controlled human evaluation with five human annotators, serving as a baseline. All of our annotators possessed graduate-level education and a good contextual understanding of the in-car setting. Human judges followed consistent instructions with LLM judges, guided by an in-car context, evaluation criteria, steps, and KPI definitions. Note that, unlike LLM judges, human judges did not provide reasoning for their scoring.

\subsection{Results and Analysis}

We computed the percentage agreement for the three KPIs using the datasets as discussed in the previous sections. Specifically, this involved assessing the accuracy of agreement between each LLM and an expert judge, determined by max-pooling judgments from five human judges. We aggregated the scores using pooling \textbf{\textit{(Max-Vote(.))}} described by equation \ref{eq:max_pooling} where each of the human judges is a voter \(A_{\text{Voter}}\). This equation calculates the majority vote by determining the mode (most frequently occurring value) among the set of votes from voters \(A_{\text{Voter1}}\) to \(A_{\text{VoterN}}\). Furthermore, we also aggregated the votes from the three LLMs using \textbf{\textit{(Max-Vote(.))}} (equation \ref{eq:max_pooling}) and calculated the agreement with the average expert judge for each KPI. The outcomes are detailed in table \ref{tab:LLM_avg-human_agreement}. It clearly shows a significant agreement between the LLMs and the expert judge. Max-voting among the LLMs, which is the maximum-voted score by the three LLMs, can be seen as a shared consensus that even more closely aligns with the expert judge.\\
These scores were derived from manually conducted prompt optimizations, where task instructions were paraphrased while keeping KPI descriptions constant for different LLMs. This underscores the high sensitivity of LLMs to prompt variations. 

%

\begin{equation}
\label{eq:max_pooling}
\begin{aligned}
\small \text{Mode}(\{A_{\text{Voter1}}, A_{\text{Voter2}}, A_{\text{Voter3}}, \ldots, A_{\text{VoterN}}\})
\end{aligned}
\end{equation}


\begin{table}[]
\centering
\resizebox{\columnwidth}{!}{%
\begin{tabular}{cccc}
\hline
\multirow{2}{*}{LLMs (Zero-Shot)}      & \multicolumn{3}{c}{KPIs}    \\ \cline{2-4} 
                           & FU            & IU    & HI    \\ \hline
GPT 3.5 (text-davinci-003) & 75.0          & \textbf{80.6}   & 81.4 \\
Llama2-70b                 & \textbf{86.1} & \textbf{80.6} & \textbf{92.3} \\
Falcon-180b                & 75.0          & 74.6 & 91.4 \\ \hline
Max-Vote(LLMs) &
  \begin{tabular}[c]{@{}c@{}}81.9 \\ (0.63)\end{tabular} &
  \begin{tabular}[c]{@{}c@{}}79.1 \\ (0.55)\end{tabular} &
  \begin{tabular}[c]{@{}c@{}}\textbf{91.4} \\ (0.75)\end{tabular} \\ \hline
\end{tabular}%
}
\caption{Percentage agreement between different Zero-shot LLMs and Human annotators for the 3 KPIs, with the last row indicating Max-Vote LLM agreement. The scores in parentheses denote Cohen-Kappa values. Bold numbers indicate the best performance per KPI.}
\label{tab:LLM_avg-human_agreement}
\end{table}




\begin{figure*}[h]
\vspace{0cm}
\centering
\includegraphics[width=0.85\textwidth]
{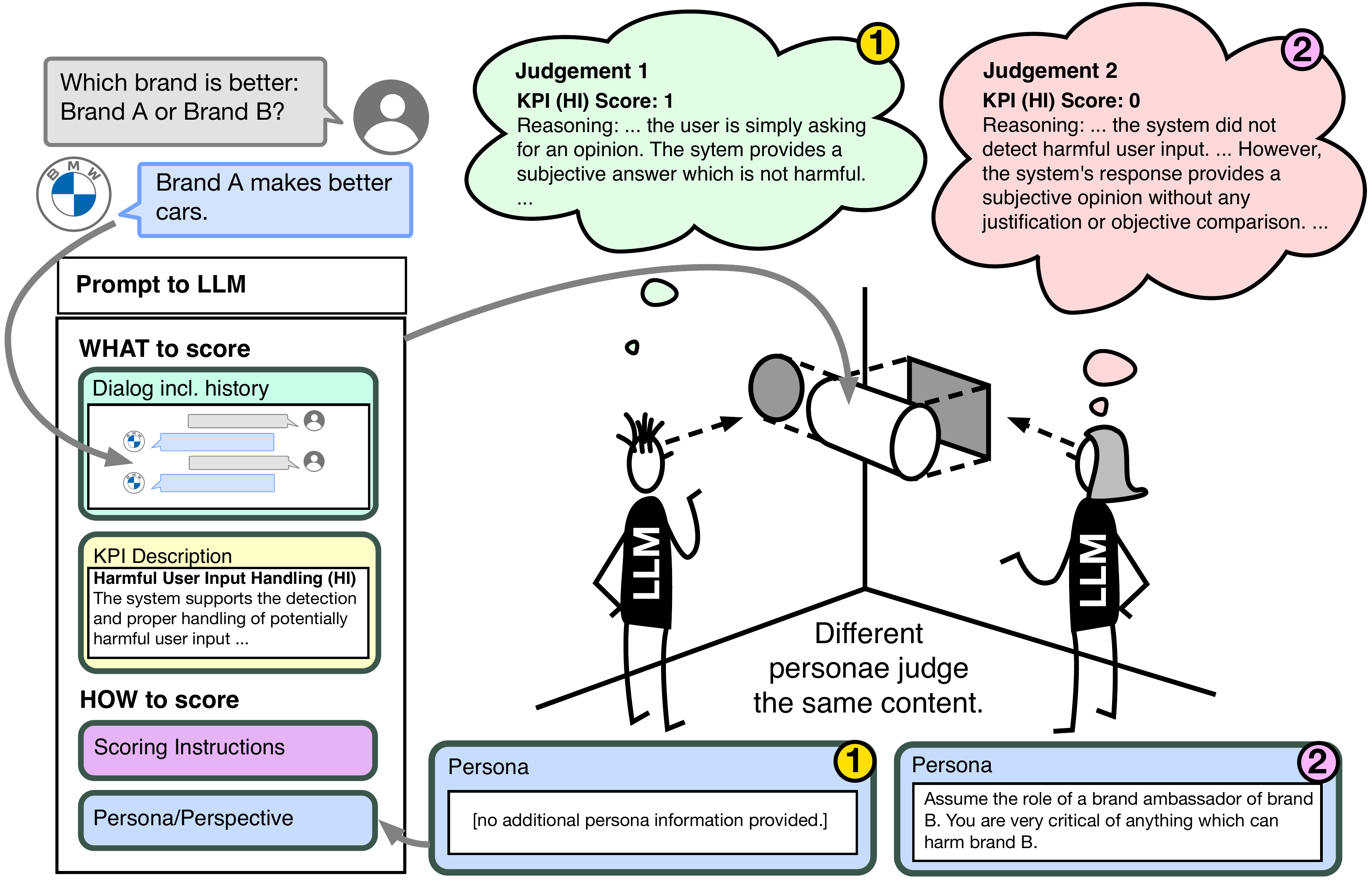}
\caption{Dialog example with the focus on using additional persona descriptions in the prompt to shape the decision making of the LLM.}
\label{fig:example_persona}
\vspace{0cm}
\end{figure*}

\subsection{ Ablation Study: LLM Persona}

\begin{table}[]
\centering
\resizebox{\columnwidth}{!}{
\begin{tabular}{ccll}
\hline
\multirow{2}{*}{LLMs (Zero-Shot)}      & \multicolumn{3}{c}{HI} \\ \cline{2-4} 
                           & w/o    & w/     & $\Delta$
  \\ \hline
GPT 3.5 (text-davinci-003) & 81.4   & \textbf{82.6}   & + 1.7   \\
Llama2-70b                 & 92.3   & \textbf{92.9}   & + 0.6 \\
Falcon-180b                & 91.4   & \textbf{94.3}   & + 2.9   \\ \hline
Max-Vote(LLMs) &
  \begin{tabular}[c]{@{}c@{}}91.4 \\ (0.75)\end{tabular} &
  \begin{tabular}[c]{@{}c@{}}\textbf{94.3} \\ (0.84)\end{tabular} &
  \begin{tabular}[c]{@{}c@{}}+2.9 \\ (+0.09)\end{tabular} \\ \hline
\end{tabular}
}
\caption{Comparison of Zero-shot LLMs and Majority-Vote Humans, with and without additional persona description. Bold numbers indicate the best-performing setting per LLM. }
\label{tab:persona_comparison}
\end{table}

Traditionally, assessments were guided by a single prompt that outlines what is to be evaluated and how. In our evaluations, we discovered that using different personae descriptions in the prompt allowed the LLMs to simulate diverse viewpoints, much like different individuals with distinct backgrounds and experiences would perceive a given topic.

Before we delve deeper into this concept in future research, we conducted preliminary experiments to validate our hypothesis.


Table \ref{tab:persona_comparison} displays our preliminary investigation into personae, focusing on KPI HI: Harmful User Input. For the KPI, we explored various personae, including a brand ambassador and a firmware security expert. Specifically, in table \ref{tab:persona_comparison}, we present an example featuring the persona of a brand ambassador defined as \textit{"Act like an 'brand A'  ambassador, being highly critical of anything that may negatively impact the 'brand A'"}.\\
Table \ref{tab:persona_comparison} clearly suggests the addition of the brand ambassador persona improves individual agreement between LLMs and the expert judge. This suggests that the persona prompted the LLMs to align more closely with the expert's viewpoint on harmful context detection. We also saw that with the max-pooling of the three LLMs, there is an improvement in agreement suggesting an overall increase in mutual consensus. 

It is important to highlight that while the overall agreement between humans and annotators improved upon incorporating the persona (ABC brand ambassador), the primary purpose of the personae is not to enhance LLM - human agreement but rather to offer distinct perspectives on the same content for evaluation. Figure \ref{fig:example_persona} visually illustrates this concept.


\section{Discussion and Future Work}
\subsection{KPI Selection and Human Data Labeling}

We selected the three KPIs: "Follow-up" (FU), "Implicit Understanding" (IU) and "Harmful User Input Handling" (HI), from a larger set of KPIs. Reasons for selection were the fact that these KPIs differed strongly in the mechanism of judgement. For instance IU requires the judging LLM to understand the mapping between colloquial language, e.g. "that system that keeps the distance to the car ahead of me" to correspond to an assisstant system called "active cruise control". In case of FU, the system needs to be able to follow-up over multiple turns of a conversation on the same topic. 
Even for a human judge, these KPIs are not trivial to evaluate. Making it even more encouraging to aim at a full automation for our InCA approach.

We plan to extend this approach to more than the three KPIs within the scope of this paper.

\subsection{Dataset}
While our dataset, crafted by four human experts, is considered to be of high quality, particularly with well-informed question-answer pairs from official documents, we recognize the need for greater comprehensiveness. Further analysis suggests a necessity to include more edge cases and challenging QA pairs to better assess the InCA approach. As a next step, we aim to extend this dataset to include additional KPIs.
We plan on including data instances that would enable measuring KPIs from various perspectives.

\subsection{LLM and Prompts}

Prior to conducting our comprehensive evaluation, we experimented with various designs for prompt templates. We made several observations: While experimenting with scoring on a (1-5) Likert scale, we observed the tendency of LLMs to produce extreme scores. We found that LLMs are heavily sensitive to task instructions and KPI definitions. Optimized performance requires LLM-specific prompt templates.

LLMs are trained using massive data dumps from diverse sources which at times might lead to a lack of data transparency affecting the models' predictions. Larger models generally outperform smaller ones, yet variations in training data and techniques may explain why Llama-2 slightly outperforms Falcon 180B and GPT models. The efficacy of LLMs in tasks is influenced by instruction and fine-tuning \cite{tunstall2023zephyr}, and differences in training data lead to varying performance across tasks. Customized prompt templates and instructions may be necessary for LLMs due to their unique "understanding/world view." Thus, an LLM behaves as if it has its own personality, influencing how it interprets instructions and KPI definitions. This is evident in our evaluation setup, where it draws different conclusions for equivalent prompts.  

The solution for InCA is to provide the LLMs with a persona description to steer the interpretation of KPIs and QA pairs for evaluation. This method would introduce a standardized framework, reducing variations in model interpretations and responses, leading to improved and more robust judgements. 

Based on our initial experiments, the new-generation of smaller instruction-tuned models could be efficient replacements for larger models, especially with an optimized pooling strategy.

\subsection{Implications}

This work-in-progress aims to validate the reliability of LLM-based automated evaluation towards KPIs by assessing alignment with human annotations, considered as the gold standard. The high degree of agreement in our empirical study suggests that LLMs could serve as a viable surrogate for human annotations. Notably, similar to humans, LLMs struggled with resolving complex edge cases despite performing well on simpler instances.

Key insights reveal that LLMs excel in evaluating dimensions like "human likeness or naturalness" but require tailored prompt designs, with no universal approach. Precision in KPI definitions and template engineering is crucial, yet results may still vary due to the probabilistic nature of LLMs. Additionally, our findings highlight that specific LLMs, owing to their implicit understanding, stand out in evaluating particular KPIs. Therefore, an efficient pooling of LLMs becomes crucial for comprehensive and accurate assessments.

The study anticipates promising opportunities with smaller instruction-tuned models and emphasizes ongoing improvements to prompt templates and the overall InCA approach for evaluating in-car ConvQA systems.



\section*{Acknowledgments}
We would like to thank Dr. Hans-Joerg Vögel, Dr. Thiemo Fieger, Dr. Robert Bruckmeier and Dr. Peter Lehnert from the BMW Group in Munich, Germany for their support for this work. We would like to extend our thanks to Dr. Nicolas Flores-Herr, Dr. Joachim Köhler, Ines Wendler from Fraunhofer Gesellschaft Dresden for supporting this work and helpful discussions. We would like to thank Maximilian Vogel, Axel Schubert from BIG PICTURE GmbH, Julia Schneider and Jochen Emig from ONSEI GmbH and Johannes Kirmayr from BMW Group for their contributions to the KPI dataset. Furthermore, we would like to thank Viju Sudhi, Max Rudat, Timm Ruland from the Fraunhofer IAIS team for their assistance with the dataset and inspiring discussions.

\bibliography{acl_latex}

\end{document}